**Edge Computing-Enabled Road Condition Monitoring: System Development and Evaluation**

**Abdulateef Daud**
Master's Student
Department of Civil and Environmental Engineering
University of Missouri-Columbia, Columbia, MO, USA, 65201
Email: aadcvg@umsystem.edu

**Mark Amo-Boateng**
Post Doctoral Fellow
Department of Civil and Environmental Engineering
University of Missouri-Columbia, Columbia, MO, USA, 65201
Email: marbz@umsystem.edu

**Neema Jakisa Owor**
Master's Student
Department of Civil and Environmental Engineering
University of Missouri-Columbia, Columbia, MO, USA, 65201
Email: nodyv@umsystem.edu

**Armstrong Aboah**
Assistant Professor
Department of Civil and Architectural Engineering and Mechanics
University of Arizona, Tucson, AZ, USA, 85721
Email: aaboah@arizona.edu

**Yaw Adu-Gyamfi**
Associate Professor
Department of Civil and Environmental Engineering
University of Missouri-Columbia, Columbia, MO, USA, 65201
Email: adugyamfi@missouri.edu

Submitted for consideration for presentation at the 103rd Annual Meeting of the Transportation Research Board, January 2024

Word Count: 7,000 words + 2 table (250 words per table) = 7,500 words

*Submitted August 1, 2023*

## ABSTRACT

Real-time pavement condition monitoring provides highway agencies with timely and accurate information that could form the basis of pavement maintenance and rehabilitation policies. Existing technologies rely heavily on manual data processing, are expensive and therefore, difficult to scale for frequent, network-level pavement condition monitoring. Additionally, these systems require sending large packets of data to the cloud which requires large storage space, are computationally expensive to process, and results in high latency. The current study proposes a solution that capitalizes on the widespread availability of affordable Micro Electro-Mechanical System (MEMS) sensors, edge computing and internet connection capabilities of microcontrollers, and deployable machine learning (ML) models to (a) design an Internet of Things (IoT)-enabled device that can be mounted on axles of vehicles to stream live pavement condition data (b) reduce latency through on-device processing and analytics of pavement condition sensor data before sending to the cloud servers. In this study, three ML models including Random Forest, LightGBM and XGBoost were trained to predict International Roughness Index (IRI) at every 0.1-mile segment. XGBoost had the highest accuracy with an RMSE and MAPE of 16.89in/mi and 20.3%, respectively. In terms of the ability to classify the IRI of pavement segments based on ride quality according to MAP-21 criteria, our proposed device achieved an average accuracy of 96.76% on I-70EB and 63.15% on South Providence. Overall, our proposed device demonstrates significant potential in providing real-time pavement condition data to State Highway Agencies (SHA) and Department of Transportation (DOTs) with a satisfactory level of accuracy.

## INTRODUCTION

As mandated by the Moving Ahead for Progress in the 21st Century (MAP-21) Act, State Departments of Transportation (DOTs) are obligated to provide accurate data regarding the condition of interstate pavements annually and non-interstate pavements biennially (*1*). However, the current data acquisition systems used by most DOTs for pavement condition data collection are very expensive to acquire, use and maintain. Consequently, the cost of continuous network-level road condition data collection is prohibitively high and impractical for most DOTs. For instance, the Virginia Department of Transportation (VDOT) spends $1.8 million per year on contractors to collect pavement roughness data using specialized profilers that have advanced sensors, lasers, and cameras (*2*). This limits their ability to perform continuous data collection and evaluation. Hence, this research aims to develop a low-cost data acquisition system that utilizes Internet of Things (IoT) embedded sensors, edge-computing, and machine learning to create a lightweight, scalable, and easily configurable device for continuous pavement condition data collection.

Prior research has utilized embedded sensors, such as triaxial accelerometers, present in contemporary smartphones as a surrogate to estimate the International Roughness Index (IRI) of pavement (*3–6*). Nevertheless, such studies have been hampered by their inability to be implemented in real-time, owing to their manual data preprocessing approaches and over dependence on cloud storage and cloud computations. In light of this challenge, there is a growing need for technologies that are less reliant on cloud storage and are capable of making on-the-fly predictions (*4*). Recent advances in internet of things (IoT) devices and edge computing have made this objective realizable. IoT devices are inexpensive, diminutive sensors and microcontrollers that can be situated on pavement surfaces or mounted on vehicles to gather insightful data on road conditions, while edge computing pertains to the processing of data at the device level, thereby circumventing the need for cloud-based computation. Our ability to combine IoT devices and edge computing will enable real-time prediction of International Roughness Index (IRI) for prompt pavement assessment.

Cloud computing has significantly revolutionized our way of life since its inception in 2006 (*7*). However, with the ever-increasing data generated by our ubiquitous end-user devices, cloud computing is ladened with two major bottlenecks – transmission latency and bandwidth requirements (*8*). Edge computing paradigm has promised to resolve these bottlenecks. The rationale behind edge computing is that computations can be carried out as close to the data source as possible. In the domain of smart cities, for instance, edge computing has been successfully implemented to address the challenges of latency and bandwidth issues. The requirement of smart cities' infrastructure is that they must be intelligent, humanized and able to make data-informed decisions in real time. Meeting this requirement may be difficult if data has to be transmitted to a remote server for computations and processing. Furthermore, edge computing has been used in traffic management systems to process traffic data collected from sensors in real-time to make timely decisions to optimize traffic flow and prevent congestion.

In field of autonomous vehicles, edge computing has been used to address the challenges of latency, safety and privacy issues. The multitude of technologies including LiDAR, GNS/IMU, Camera, Radar and Sonar systems connected together in Autonomous vehicle generate extremely large amount of data. About 1 gigabyte of data is generated per second by Autonomous vehicles (*9*). Transmitting this quantity of data to a central cloud for processing and analytics can cause delays and interruptions which can impede real-time functionality required in Autonomous vehicles. Moreover, depending on a centralized data center for decision making such critical decision is quite risky, as it requires a reliable and uninterrupted network connection, which may be lacking in remote and rural communities. Edge computing allows data to be processed locally in the vehicle, reducing latency and ensuring reliable decision-making. That is, edge computing is currently being used to process sensor data from autonomous vehicles in real-time to make immediate decisions to avoid collisions and ensure safe driving.

As previously explained, edge computing has demonstrated successful implementation in various domains, including smart cities, traffic management, and autonomous vehicles, by addressing the challenges of cloud computing. However, its use in real-time pavement monitoring is limited. As such, the current study describes the development of a low-cost pavement condition monitoring system using edge computing, which is a promising application of this technology in the transportation domain, specifically pavement condition monitoring.

The main goal of the study is to develop a low-cost embedded system that leverages IoT embedded sensors, edge-computing, and machine learning for real-time pavement condition monitoring. To achieve this goal, this study formulated 3 main objectives. The study's objectives can be summarized as follows:

- **Develop a cost-effective pavement condition data collection device**: The study aimed to design a universal, axle-based pavement roughness data collection system using tri-axial accelerometer sensors. This device, built on an ESP32 microcontroller unit, addressed the limitations of expensive data collection methods.
- **Enable fast and real-time prediction of International Roughness Index (IRI):** The study implemented edge computing techniques to process the collected data in real-time. By utilizing Fast Fourier Transform (FFT) and extracting important features from the acceleration signal, the study aimed to reduce latency and bandwidth usage, enabling efficient and timely IRI predictions.
- **Perform a comparative analysis of different models**: The third objective is to evaluate and compare the performance of three machine learning models including Random Forest, XGBoost and LightGBM in predicting IRI based on some defined evaluation metrics.

Overall, the study aimed to develop a cost-effective data collection device, enable fast and real-time IRI prediction through edge computing, and conduct a comparative analysis of different models to enhance the understanding and accuracy of IRI predictions.

## LITERATURE REVIEW
### Classical Pavement Roughness Evaluation and Data Collection Approaches
There are many indices used for evaluating pavement roughness, including the pavement condition index (PCI), pavement serviceability rating (PSR), pavement surface evaluation and rating (PASER), and International Roughness Index (IRI), among others (*10*). The U.S. Army Corps of Engineers developed PCI as an index for pavement system rehabilitation, maintenance, and overall management. PCI has values ranging between 0 and 100, with 0 indicating pavement in the worst condition possible and 100 meaning a newly paved road (*11*). On the other hand, PASER, developed by the University of Wisconsin-Madison Transportation Information Center, rates pavement on a scale of 1 to 10 based on visual inspection (*12*). PASER ratings do not require pavement distress information. This makes it suitable for instantaneous rating of road segments. One critical limitation of PASER is that it is not ideal for evaluating pavement based on distress types which is crucial in mechanistic-empirical pavement design (*12*). International Roughness Index (IRI) developed by the world bank in 1986 from the International Roughness Index Experiment (IRIE) is arguably the most widely used pavement roughness condition index. This is because it established a common standard measure of roughness adaptable to other methods by which road roughness can be characterized. IRI is mathematically formulated to depict how a single tire on a vehicle suspension (quarter-car) responds to unevenness on the road surface (as depicted in **Figure 1**), while traveling at a speed of 50 mph (*13*). The algorithm for simulating the quarter-car is extensively described in the appendix section of ASTM E 1364 (*14*). Originally, the IRI experiment was designed to conform to two road roughness measurement methods: profile-based and Response-type Road roughness measuring systems (RTRRMSs) (*13*).

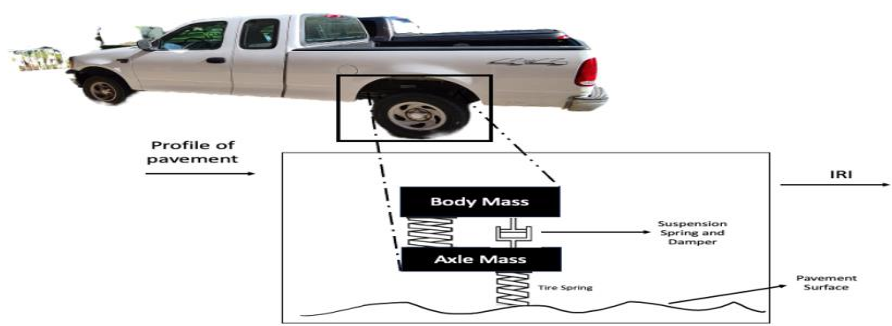

**Figure 1 Quarter Car Model Simulation (*15*)**

The profile-based method directly measures road roughness on a standard scale obtained from longitudinal road profiles. They range from simple Dipstick profilers to more sophisticated high-speed profilometers. Although these high-precision profilers provide timely and very accurate IRI readings, they have the disadvantage of being prohibitively expensive to acquire, operate and maintain. RTRRMSs, on the other hand, give IRI estimates from regression equations used to calibrate the readings from RTRRMSs. Unlike profilers that measure the road roughness directly, RTRRMSs rely on the measuring vehicle's roughness-induced dynamics (vertical accelerations, displacement) to estimate IRI (*13*). They are cheaper, easier to operate, and highly scalable; however, this comes at the cost of accuracy as they are highly susceptible to environmental changes and highly system-dependent.

### Digital Signal Processing and ML/DL Modeling Approaches
A smartphone-based and cloud-assisted road condition monitoring platform was implemented by (*16*), taking advantage of sensors present in mobile phones and cloud servers' computational power. The framework can classify road conditions as either rough, average, or smooth based on the accelerometer readings and issue warnings accordingly. However, this road roughness classification is too coarse and not quantitative enough for state DOTs. The result from the study confirms that real-time classification and monitoring of road conditions is possible as the lightweight mobile application can classify road conditions at an accuracy of 88.6% within 1.3 milliseconds. Despite this performance, this study is limited in several ways. One of these limitations is the study's overreliance on accelerometer data. Vehicle speed, vehicle type, suspension system condition, etc., are some of the many other relevant information that could affect the accelerometer data obtained from the sensors.

In another study conducted by (*3*), the authors proposed a deep learning framework with entity embeddings for accurate and timely prediction of IRI based on smartphone sensor data and historical IRI data. Due to the high cost of existing data collection methods, this study developed a smartphone app for road surface data collection. Unlike (*16*), who solely rely vertical acceleration data, this study included speed and gyroscope data for more accurate IRI estimation. First, the study used an Empirical Mode Decomposition algorithm – a non-stationary signal decomposition algorithm to resolve the complex signal into different components to extract features important in characterizing pavement roughness. Next, Fourier Transform was applied to transform the decomposed signals to a power-frequency domain (power spectral density). The total area under the power spectral density, maximum power, and the corresponding frequency on each resolved signal and historical IRI are input features to a convolution neural network (CNN). When evaluated on the test data, the model's average root means square error (ARMSE) and root-mean square-percent-error (RMSPE) were 5.5 and 0.17, respectively. One critical limitation of the study is that the data preprocessing approach used in this study was manual and therefore not suitable for real-time implementations.

## Edge Computing and Internet of Things (IoT) for Pavement Condition Monitoring

Recent studies have applied the concept of edge computing in Transportation Systems. In a study conducted by (*17*), the authors identified the limitations of cloud computing in meeting the low latency and context awareness requirements of Transportation Cyber-Physical Systems (TCPS). To solve this problem, the study proposed Mobile Edge Computing (MEC) as a solution. However, MEC has limitations in processing computationally intensive tasks like deep learning algorithms. To address this, a lightweight deep learning model called lightweight CNN-FC was developed to support MEC applications in T-CPS. The proposed model reduces unnecessary parameters and decreases model size while maintaining high accuracy compared to conventional CNN models. Experimental results on a realistic MEC platform (Jetson TX2) show that the proposed model is quite accurate and portable. In another study, (*18*) designed and evaluated an edge-computing device using computer vision and deep neural networks for real-time tracking in multi-modal transit. Building on this foundation, (*19*) applied edge computing using low-cost Internet of Things (IoT) devices and Long-Range Wide Area Network (LoRaWAN) to count the number of vehicles and report traffic counts to the servers. The author trained and compressed a YOLOv3 object detection model to make it deployable on the Raspberry Pi. The result from this experimental work demonstrates the effectiveness of the proposed approach in real-world vehicle detection and counting using edge-based video analytics.

## METHODS

### Problem Formulation

The goal of this study is to predict the International Roughness Index (IRI) for every 1/10th of a mile ($d_{thr}$) using the vertical acceleration ($a_z$) and GPS data obtained from our proposed device as input features. The approach involves using a supervised machine learning model that takes the primary features derived from the GPS data and the secondary features extracted from the vertical acceleration data transformed through the Fast Fourier Transform (FFT) algorithm. Specifically, given that $IRI \in \mathbb{R}^n$ represent the ground truth IRI for $n$ 1/10th mile intervals and $X \in \mathbb{R}^{n \times 2}$ be the primary features matrix, where each row $x_i = \{speed_i, \, elevation_i\}^T$ represents the aggregated speed and elevation data obtained from our proposed device at every $d_{thr}$. Let $d_{thr}$ be the secondary features matrix, where each row $y_i = \{auc_i, \, mp_i, \, sdp_i, \, mxp_i, \, df_i\}^T$ represents the features extracted from the vertical acceleration data at every $d_{thr}$. The features include the area under the Power Spectral Density (PSD) curve ($auc_i$), mean power ($mp_i$), standard deviation of power ($sdp_i$), maximum power ($mxp_i$), and dominant frequency ($df_i$). The FFT algorithm is used to transform the vertical acceleration data from the time domain to the frequency domain as shown in **Figure 2**. Mathematically, given a discrete-time signal $a(n)$ of length N, where $n = 0,1,2,\ldots,N-1$, the FFT computes the Discrete Fourier Transform (DFT) of the signal, which is denoted as $A_{(K)}$ for $k = 0,1,2,\ldots,N-1$. The transformation equation is $A_{(K)} = \sum_{n=0}^{N-1} a(n) \, e^{-i2\pi kn/N}$ .

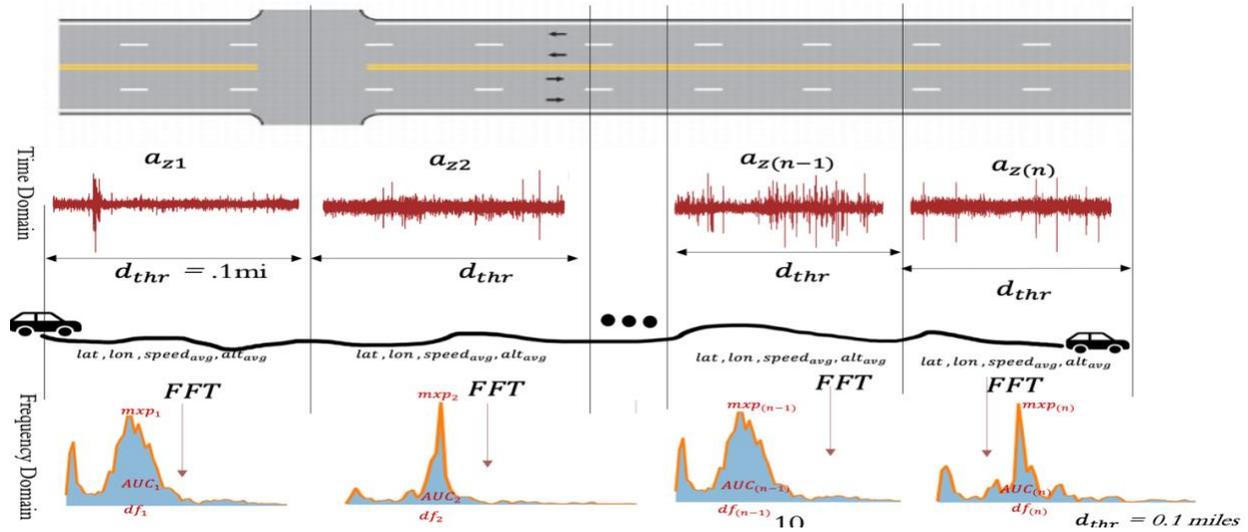

**Figure 2 Conceptual Diagram of the Formulated Problem**

In the context of this study, the vertical acceleration data ($a_z$) at each $1/10th$ mile ($d_{thr}$) is represented as a time-domain signal. The FFT algorithm is applied to this signal to obtain the frequency-domain representation, providing insights into the frequency components present in the data. The supervised machine learning model, denoted as $f(X, Y)$, aims to predict the IRI values for $1/10th$ mile ($d_{thr}$) interval. Mathematically, this can be represented as $IRI = f(X, Y)$ where $f \colon \mathbb{R}^{n \times 2} \times \mathbb{R}^{n \times 5} \to \mathbb{R}^n$ is the function that maps the combined primary features X and secondary features Y to the predicted IRI values. By combining the aggregated GPS data as primary features and the secondary features extracted from the vertical acceleration ($a_z$) data using the FFT algorithm, the machine learning model can be trained to predict the IRI values for each $1/10th$ mile ($d_{thr}$). This approach leverages the speed, elevation, and spectral characteristics of the data to estimate road surface roughness based on the provided features.

### Proposed Methodology

The general methodology employed in this study for real-time IRI prediction comprises of four distinct phases as depicted in **Figure 3**. At the data collection phase, the proposed device was configured to collect vertical acceleration ($a_z$) and GPS data as the data collection vehicle traverses the route of interest. Next, the vertical acceleration signal corresponding to every $d_{thr}$ segment was passed through Fast Fourier Transform (FFT) to convert it from the time-domain to the frequency domain. This transformation facilitated the extraction of secondary features during the data preprocessing stage. The speed and altitude from the GPS were also aggregated at every $d_{thr}$. The distance in miles was estimated from the GPS coordinate using the haversine formulae. At the modeling and prediction phase, the aggregated data alongside the MoDOT's IRI serving as the ground truths (labels), were utilized as inputs to train supervised machine learning models. These trained models are expected to map newly collected accelerometer and GPS data to their corresponding IRI at every $d_{thr}$. Finally, during the deployment phase, the data preprocessing algorithms and the trained machine learning model were deployed on an edge computing device. This enabled the device to collect data, instantly preprocess and aggregate the data on the edge, and employ the deployed machine learning model to predict the IRI in real-time.

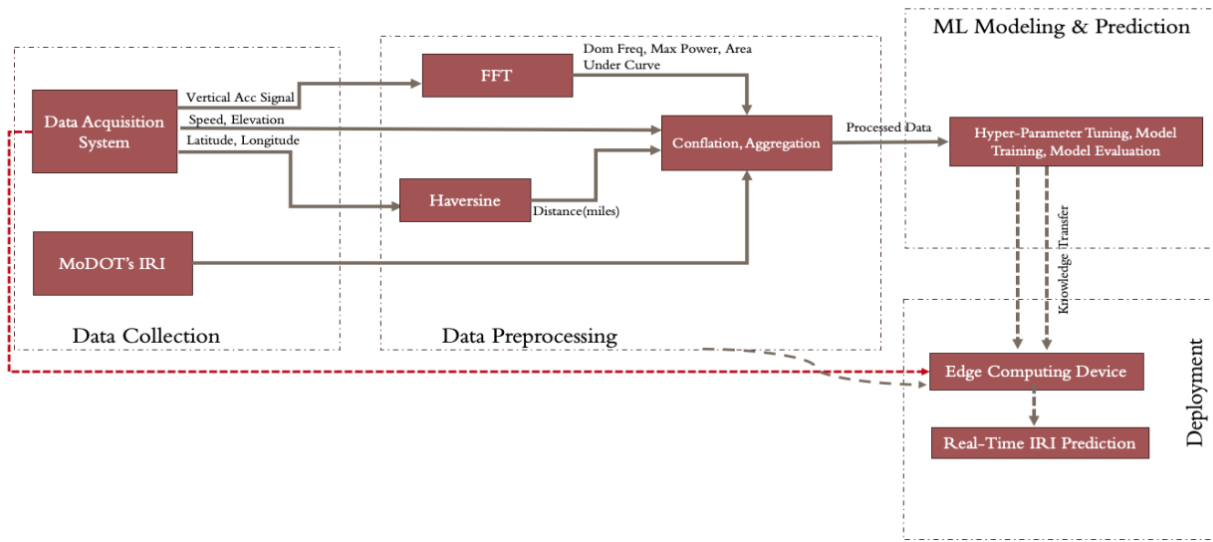

**Figure 3 Flowchart of Proposed Methodology**

*Ensemble Machine Learning Models*

Ensemble machine learning — a technique that combines several base models (learners) to produce a single optimal model — have been reported in literature to produce superior performance than traditional machine learning models. Bootstrap Aggregation (Bagging) and Boosting are two of the popular approaches used in ensemble machine learning models. In Bagging, data subsets are generated by randomly sampling the original dataset with replacement, each data subsets are then used to train separate base models in a parallel manner as shown in **Figure 4** (a), the predictions from the models are combined to produce a final prediction. Boosting, on the other hand, involves sequentially training each base model on a randomly sampled datasets such that each model compensates for the weakness of its predecessor as shown in **Figure 4**(b). This ultimately convert a set of weak learners (base model) into a strong learner with better performance. In this study, we utilized three ensemble machine learning models; Random Forest model (a variant of the bagging method), extreme Gradient Boosting (XGBoost) and Light Gradient-Boosting Machine (LightGBM) (variants of the Boosting method).

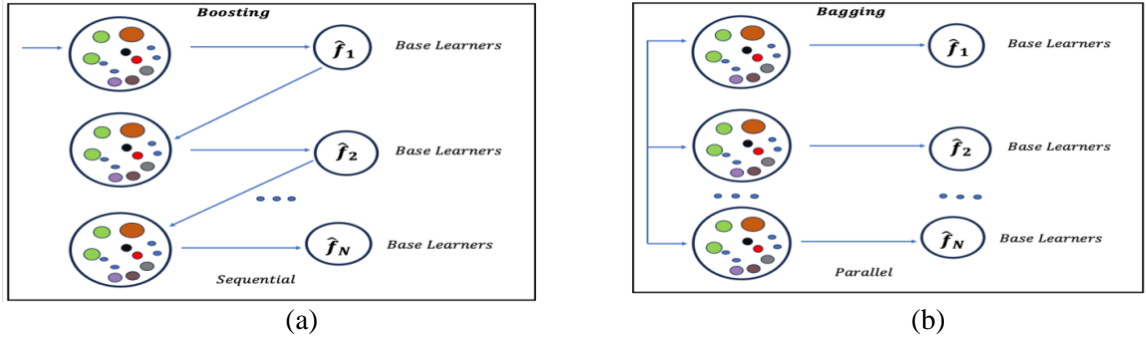

**Figure 4 (a) Boosting-Based Ensemble Method (b) Bagging-Based Ensemble Method**

### Data Preprocessing and FFT-based Feature Extraction

The data preprocessing approach used in this study is illustrated in **Table 1**. The first step is to estimate the distances from the GPS coordinates for all the data instances using the haversine formula. Then the vertical acceleration data ($a_z$) from the accelerometer (MPU6050), the speed ($spd$) and altitude ($alt$) from the GPS are then grouped based on a set distance threshold $0.1\ mile(d_{thr})$. This means that the $a_z$, $spd$ and $alt$ will be grouped together until the distance threshold condition is met after which it falls to the next group. Next, the $a_z$ from each of these groups are then treated as signals to the Fast Fourier Transform that covert them from the time domain to the frequency domain. This transformation results in a Power Spectral Density curve from which spectral features such as dominant frequency, maximum power ($mxp$), area under curve ($auc$), standard deviation of power ($stp$) and the mean of power ($mp$) can be extracted. These features are very crucial in characterizing the spectra from the signals. Also, the mean of the speed and altitude for each of these groups are estimated and appended to the extracted spectral features from each of the grouped signals. These features alongside the ground truth IRI from MoDOT serve as the input to our supervised machine learning models.

**TABLE 1 FFT-Based Feature Extraction Algorithm**

| Steps | FFT-based Feature Extraction Algorithm |
|---|---|
| 1 | **Require: $a_z$, $\varphi_1, \varphi_2, \lambda_1, \lambda_2, r$:** vertical acceleration, latitude, previous latitude, |
| 2 | longitude, previous longitude, radius of earth. |
| 3 | **Require: $d_{thr}$, $spd, alt$:** distance threshold, speed, altitude |
| 4 | **Feature Extraction (N: Sample size):** |
| 5 | $i \leftarrow 0$ (Initialize counter), $f_{max} \leftarrow [\ ]$, $P_{max} \leftarrow [\ ]$, $P_{std} \leftarrow [\ ]$, |
| 6 | $P_{mean} \leftarrow [\ ]$, $AUC \leftarrow [\ ]$, $spd \leftarrow [\ ]$, $alt \leftarrow [\ ]$ |
| 7 | $d \leftarrow 2\ r\ arcsin\left(\sqrt{sin^2\left(\frac{\varphi_2 - \varphi_1}{2}\right) + cos\varphi_1 cos\varphi_2 sin^2\left(\frac{\lambda_2 - \lambda_1}{2}\right)}\right)$ |
| 8 | **While Loop $d < d_{thr}$** |
| 9 | $signal(n) \leftarrow a_z$ |
| 10 | $Power, Freq = \sum_{n=0}^{N-1} signal(n)\, e^{-i2\pi kn/N}$ |
| 11 | $speed \leftarrow speed_i$ |
| 12 | $altitude \leftarrow altitude_i$ |
| 13 | If $d = d_{thr}$, then $i = N$ |
| 14 | $P_{mean} \leftarrow mean(Power), P_{std} \leftarrow std(Power), AUC \leftarrow sum(power)$ |
| 15 | $f_{max} \leftarrow max(freq), P_{max} \leftarrow max(power), spd \leftarrow mean(speed),$ |
| 16 | $alt \leftarrow mean(speed)$ |
| 17 | Output: $P_{mean}, P_{std}, AUC, f_{max}, P_{max}, spd, alt$ |

*Edge Computing Device*

One of the objectives of this study is to design an edge computing system that collect road condition data and perform on-board preprocessing of the data, and give real time predictions of IRI based on a deployed machine learning model on the system. The device used in this study comprises of the ESP32 microprocessor, the GPS, the MPU 6050 as shown in **Figure 5**

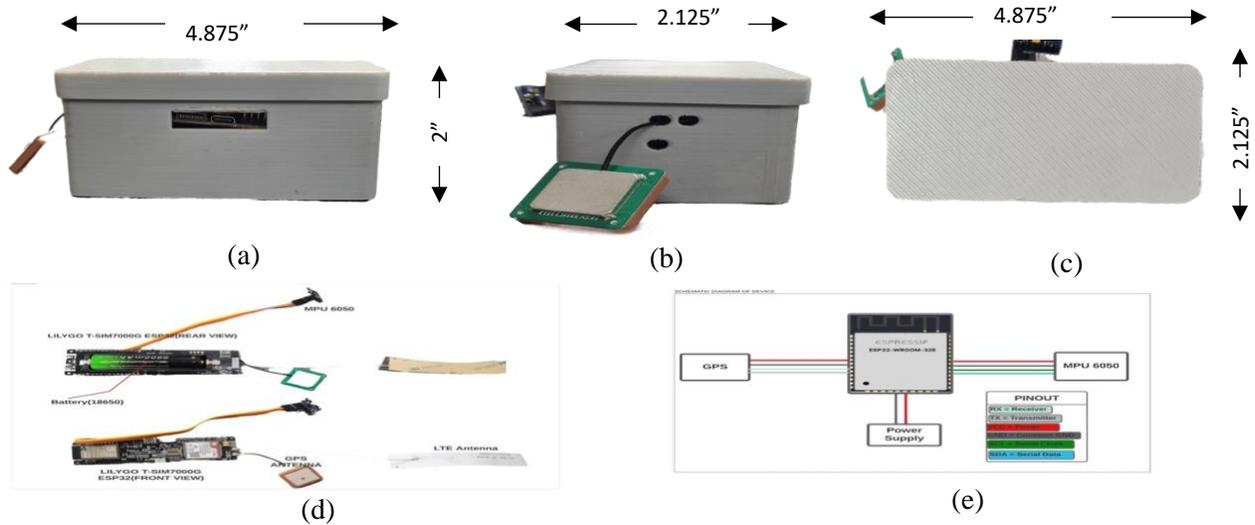

(a)      (b)      (c)

(d)      (e)

**Figure 5 Proposed Edge Computing Device: (a)Proposed Device in Enclosure (Side View) (b) Proposed Device in Enclosure (Front View) (c) Proposed Device in Enclosure (Top View) (d) Propose Device Without Enclosure (e) Schematic Diagram of Proposed Device Components**

## DATA COLLECTION

In this study, multiple streams of datasets were collected from multiple devices and sources including the sensor data from our proposed device, smartphone collected data and IRI data measured by using MoDOT's Automatic Road Analyzer (ARAN) vehicle. Our proposed device was configured, set up and mounted on the axle of the survey vehicle as shown in **Figure 6**(a). As the vehicle traverses the route of interest, our mounted device captures the vertical acceleration data (using the MPU6050 module) and GPS data at sampling rate of 365Hz. The GPS data includes the GPS coordinates, speed, altitude and timestamps. As the device captures these datasets, they are simultaneously logged to an SD card in a text file format on the device for further data processing.   On the same routes, smartphone data was also collected using the Tiger-Eye App which runs on the Apple Operating System (iOS). The set-up of the smart-phone based data collection was shown in **Figure 6**(b). The Tiger-Eye App samples accelerometer data and GPS data at a frequency of 100Hz. The smartphone also has the capability to capture the imagery which helps us to cross-check the predictions from our proposed device with the actual visual distresses observed from the imagery.

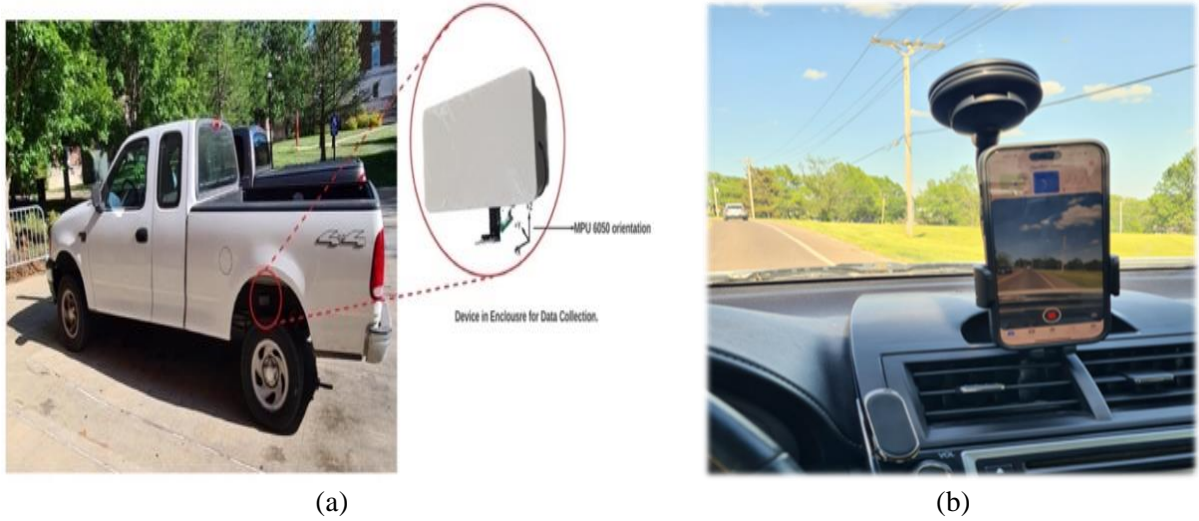

(a)                                                              (b)

**Figure 6 (a) Data Collection Set-up for our Proposed Edge Computing Device (b) Smart Phone Data Collection Set-Up Using the Tiger-Eye App on (iPhone 14 Pro Max)**

In order to validate our collected data, the IRI values measured by MoDOT's ARAN van was used as ground truth in each of our models. This data was obtained from the MoDOT's Transportation Management System's (TMS) database. The interface of the website is shown in **Figure 7**. This IRI value is reported at every 0.1 mile between the Start-Log and End-Log of each segment. This informed our aggregation approach to be based on 0.1 miles to conflate the MoDOT's reference IRI data with the data from our proposed device and the Tiger Eye App.

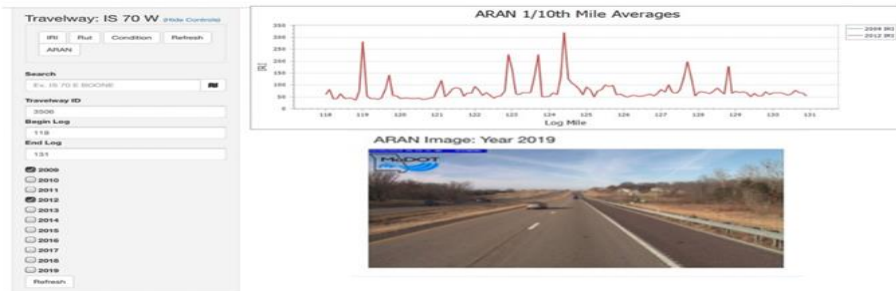

**Figure 7 Interface of the MoDOT's TMS website showing a line plot of IRI and a picture of I-70WB (*3*)**

**Selected Routes**

For this study, the routes selected are I-70EB, I-70WB, South-Providence and Route-K totaling 159.4 miles. These routes are selected based on ARAN data availability on the MoDOT's transportation management system (TMS) database. **Table 2** below shows the length, speed limits, coordinates and the names of all 4 selected routes. **Figure 8** shows a map visualization of each of the selected routes and images from the video data of each route.

**TABLE 2 TravelWay ID, Coordinates, Length, and Speed-Limit of Selected Routes**

| Device | Route Names | Travel Way ID | Start Coordinates | End-Coordinates | Length(mi) | Speed Limit (mph) |
|--------|-------------|---------------|-------------------|-----------------|------------|-------------------|
| **Proposed Device** | I-70 EB | 19 | 38.958542, -92.206479 | 38.809596, -90.878069 | 74.4 | 60-70 |
| | I-70 WB | 3506 | 38.809424, -90.879521 | 38.961691, -92.286978 | 78.6 | 60-70 |
| | South-Providence | 7259 | 38.937625, -92.334454 | 38.893154, -92.335705 | 3.2 | 50 |
| | Route-K | 3537 | 38.891846, -92.336358 | 38.870686, -92.381846 | 3.2 | 45 |
| **Tiger-Eye Mobile App** | I-70 EB | 19 | 38.958542, -92.206479 | 38.809596, -90.878069 | 74.4 | 60-70 |
| | I-70 WB | 3506 | 38.809424, -90.879521 | 38.961691, -92.286978 | 78.6 | 60-70 |
| | South-Providence | 7259 | 38.937625, -92.334454 | 38.893154, -92.335705 | 3.2 | 50 |
| | Route-K | 3537 | 38.891846, -92.336358 | 38.870686, -92.381846 | 3.2 | 45 |
| **MoDOT ARAN van** | I-70 EB | 19 | 38.958542, -92.206479 | 38.809596, -90.878069 | 74.4 | 60-70 |
| | I-70 WB | 3506 | 38.809424, -90.879521 | 38.961691, -92.286978 | 78.6 | 60-70 |
| | South-Providence | 7259 | 38.937625, -92.334454 | 38.893154, -92.335705 | 3.2 | 50 |
| | Route-K | 3537 | 38.891846, -92.336358 | 38.870686, -92.381846 | 3.2 | 45 |

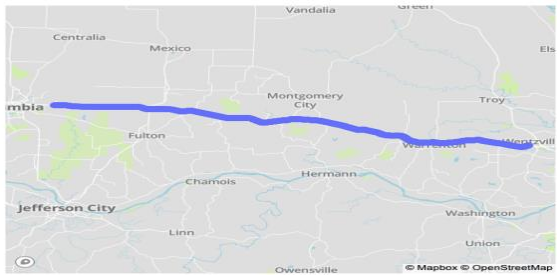

(a)

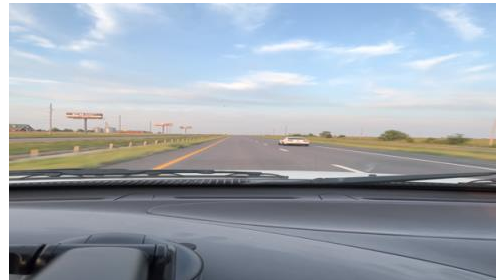

(e)

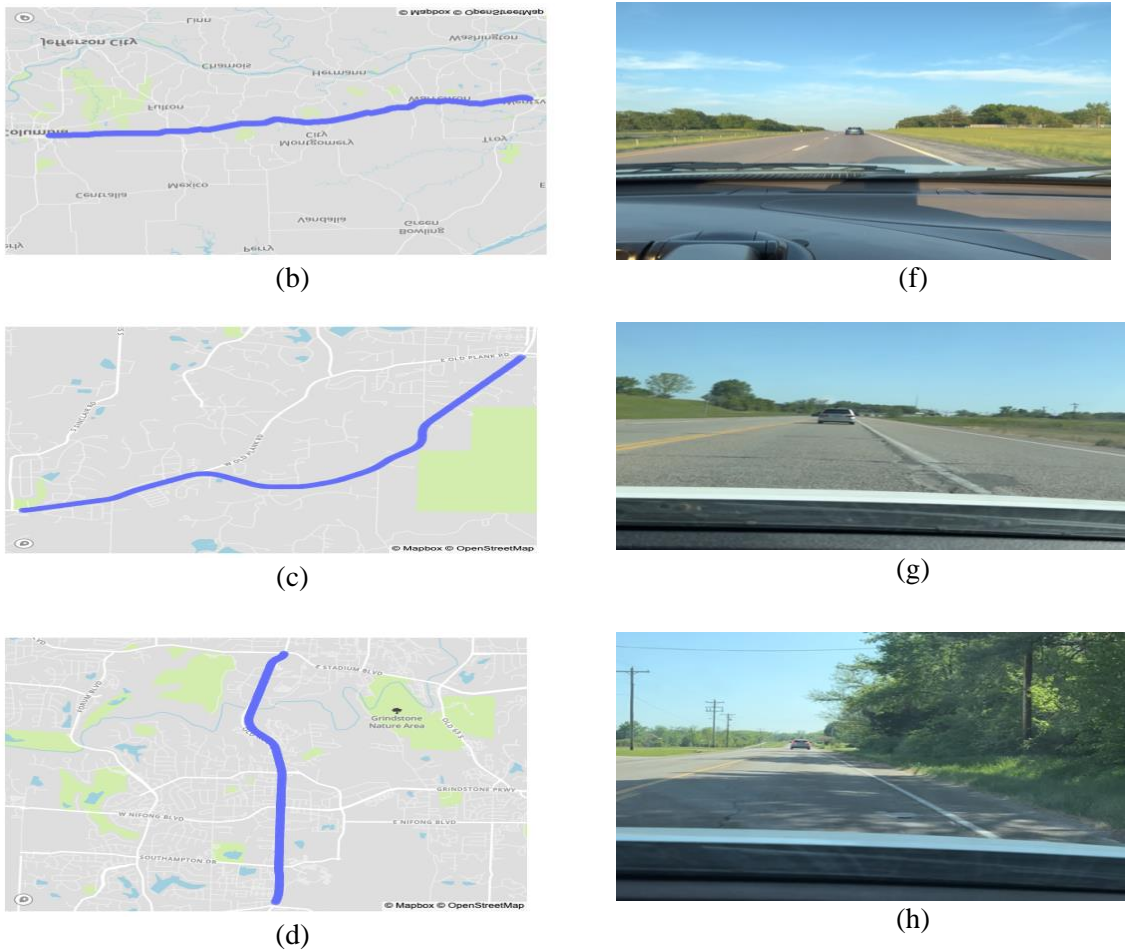

**Figure 8 Map Visualization and Captured Images of Selected Routes: (a) I-70EB Map (b)I - 70WB Map (c) South Providence Map(d) Route-K Map (e) I-70EB Image (f) I - 70WB Image (g) South Providence Image(h) Route-K Image**

## Model Evaluation Metrics

In this study, RMSE (Root Mean Square Error), MAPE (Mean Absolute Percentage Error) and $R^2$ are employed as evaluation metrics to assess the performance of the developed models. The RMSE is the square root of the mean of the squared difference between the IRI predicted by the model and the MoDOT's ground truth IRI as shown in **Equation 1**. The lower the RMSE value, the better the model's performance. MAPE estimates the mean percentage difference between the predicted IRI and the ground truth IRI as shown in **Equation 2.** It expresses the prediction error as a percentage of the actual values. This makes it easier to compare performance across different models. The lower the MAPE, the better the model's performance. $R^2$, shown in **Equation 3,** quantifies the variability in the datasets explained by the model.

$$RMSE = \sqrt{\frac{1}{n}\sum_{i=1}^{n}(y_i - \hat{y}_i)^2} \qquad (1)$$

$$MAPE = 100\% \frac{1}{n}\sum_{i=1}^{n}\left|\frac{y_i - \hat{y}_i}{y_i}\right| \qquad (2)$$

$$R^2 = 1 - \frac{\sum_{i=1}^{n}(y_i - \hat{y}_i)^2}{\sum_{i=1}^{n}(y_i - \bar{y})^2} \qquad (3)$$

Where:
$y_i$ = predicted IRI
$\hat{y}_i$= ground truth IRI
$n$= number of data points

## RESULTS AND DISCUSSIONS
### Comparative Analysis of Machine Learning Models
**Figure 9** (a) through (c) shows plots of MoDOT's ground truth IRI against predicted IRI for the Random Forest, XGBoost and LightGBM models, respectively. The horizontal axis represents the ground truth IRI and the vertical axis represent the predicted IRI. For the random forest model, it can be observed that there is an acceptable level of correlation between the predicted IRI and the ground truth IRI with $R^2$ of 0.63. The RMSE and MAPE are 16.96in/mi and 22.23%, respectively. This indicates that predictions are on average 16.96 in/mi and 22.23% off the ground truth estimate from MoDOT. For the LightGBM model, the RMSE and MAPE are 17.69in/mi and 20.87% respectively. The $R^2$ for Light GBM was 0.61 which was also a bit lower compared to that of Random Forest. XGBoost model has the best performance with an RMSE and MAPE of 16.89in/mi and 20.3%, respectively. It also has the highest correlation between the predicted IRI and the actual IRI with an $R^2$ of 0.64. These comparative results revealed that the XGBoost model outperforms the Random Forest and LightGBM models in terms of accuracy and correlation with the ground truth IRI. It demonstrates lower RMSE and MAPE values, indicating a closer match between predicted and actual IRI values.

**Figure 10** shows a line chart of the IRI predictions from all three models and the actual IRI, with the vertical axis representing the IRI and the horizontal axis represent the index. It can be observed that, overall, the IRI predictions from Random Forest, LightGBM and XGBoost models almost aligned with the trend set by the ground truth IRI. For example, at region A, B C and D the ground truth IRI and the models' predicted IRI was very close, this trend can be observed in most part of the line chart. However, some deviations between the models' prediction and the ground truth IRI can also be observed. For instance, at region E and G, the predictions from the three models were lower compared to the ground truth IRI. Specifically, at region E, the ground truth IRI is 221inches/miles, while the predicted IRI from XGBoost, LightGBM and Random Forest were 113inches/mile, 119inches/mile, 109inches/mile, respectively. Furthermore, there were cases where the models' predictions overshot the ground truth IRI, as can be observed at region F and H. At region H, the ground truth IRI was 53.54inches/mile while, while XGBoost, LightGBM and Random Forest predicted IRI were 74.67 inches/mile, 121.04inches/mile and 73.34inches/mile respectively.

The marked deviation of the model's IRI predictions from the ground truth IRI that can be observed at region E and G can be attributed to the recently completed road resurfacing that was noticed during the data collection along some segments in the selected routes. **Figure 11** shows a picture taken during the data collection along I-70WB where some segments have been resurfaced. Also, at region F and H where the predicted IRI from all the models used overshot the ground truth IRI can be justified by the fact that the most recent IRI data from MoDOTs' ARAN viewer portal was as far back as 2021. The axle-based sensor data used in this study was collected 19th December 2022. Deterioration of the road surface over time due to factors such as wear and tear between road surface and vehicle tires, repeated freezing and thawing cycles due to extreme weather the road surface has been subjected to may have caused these discrepancies. Also, it should also be pointed out that the profilers used in the collection of the ground truth IRI data are not 100% error-free.

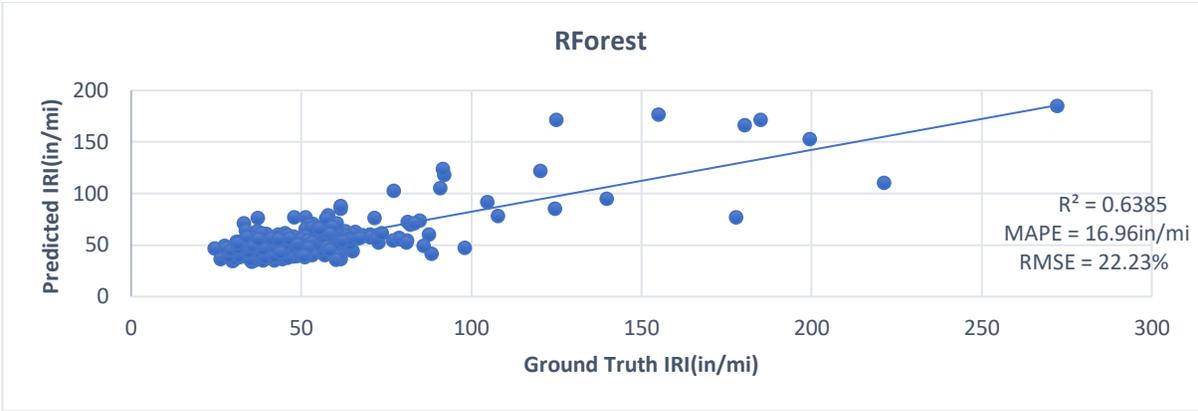

(a)

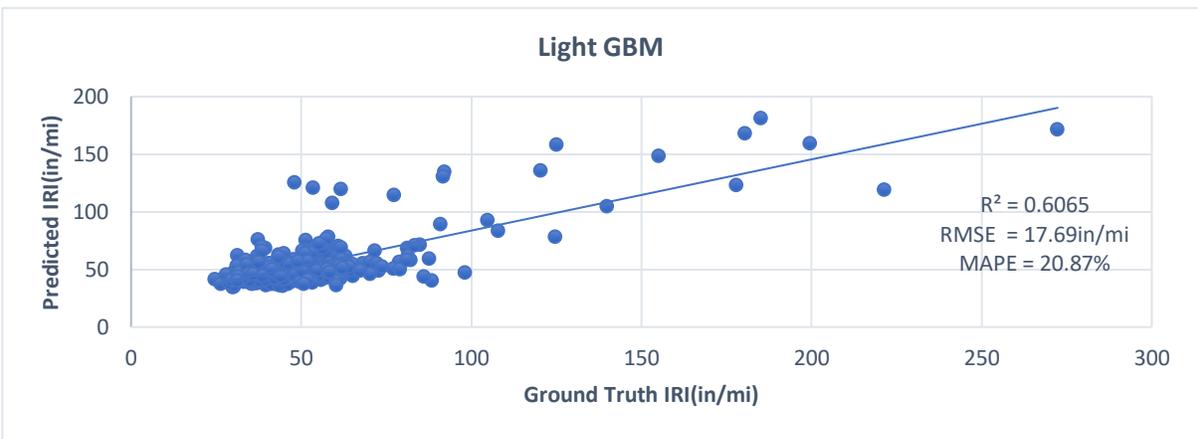

(b)

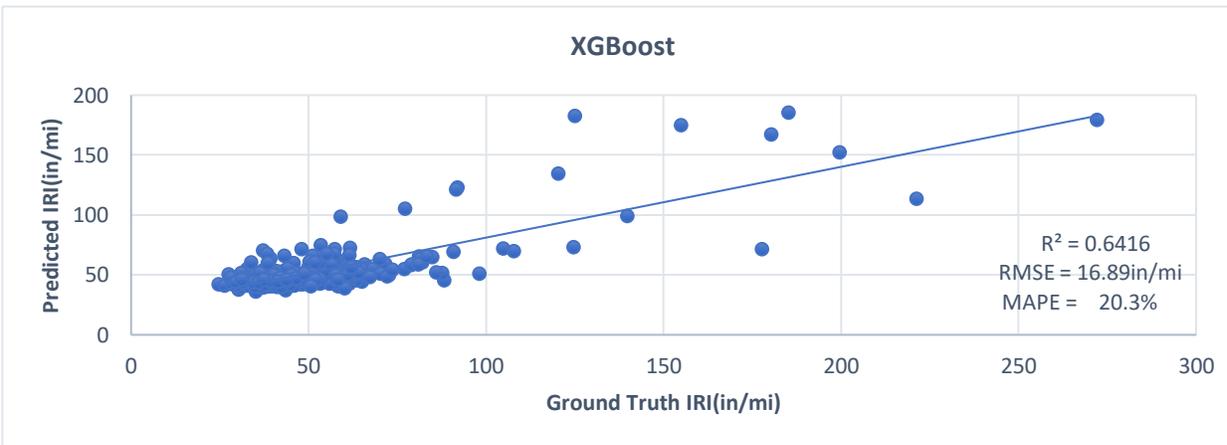

(c)

**Figure 9 Plot of Predicted IRI against Ground Truth IRI for Random Forest Model (a) Random Forest Model (b) LightGBM (c) XGBoost**

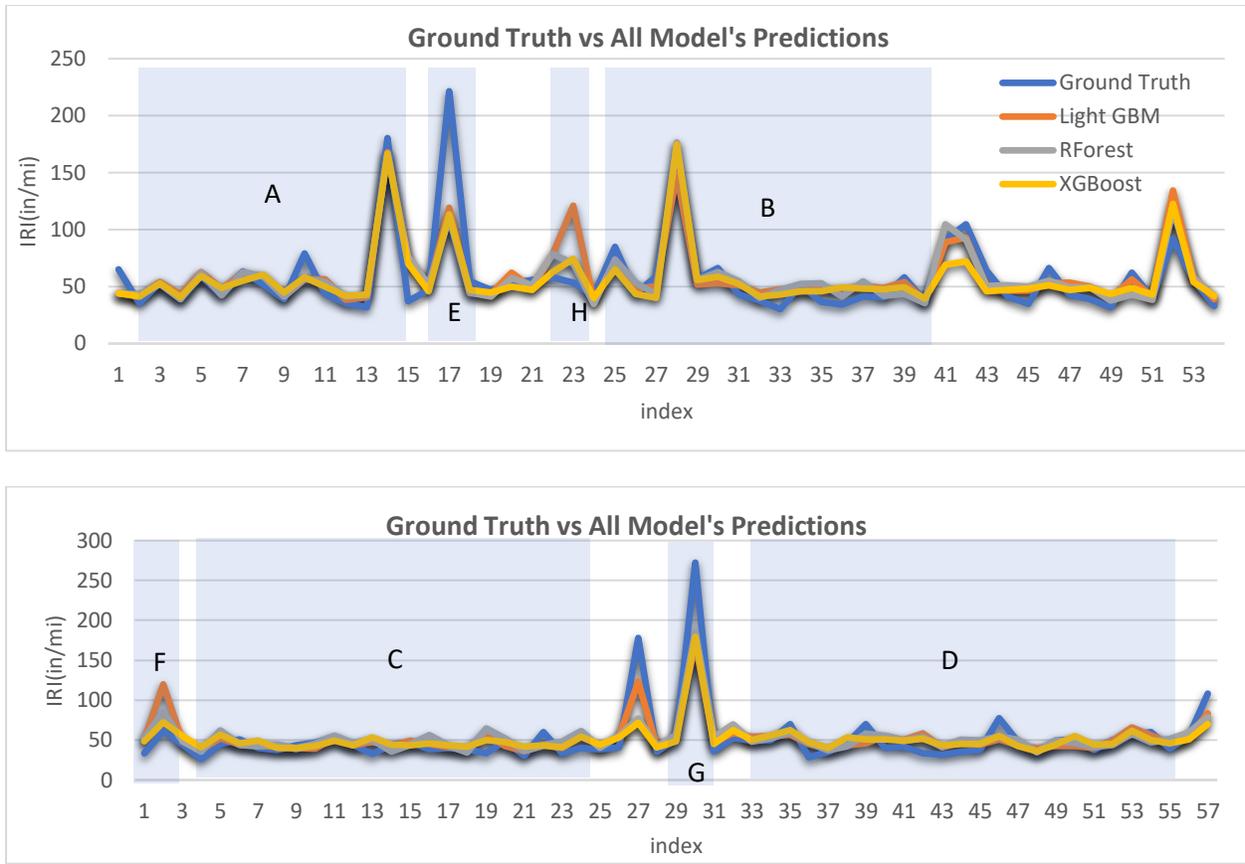

**Figure 10 Line plot of ground truth IRI and predicted IRI from Random Forest, LightGBM and XGBoost**

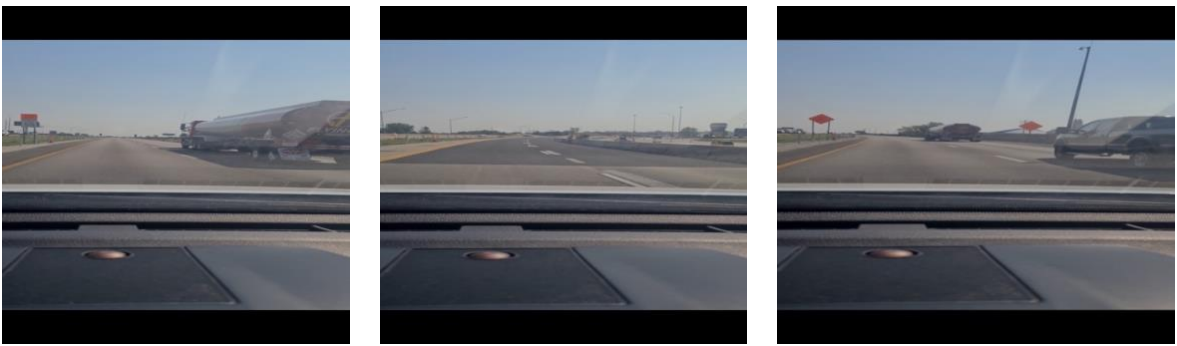

**Figure 11 Images of newly resurfaced roads on I70EB during data collection**

**Comparing Proposed Device Predictions with Ground Truth IRI for I-70EB**

**Figure 12** (a) and (b) present line plots of IRI predictions from our proposed device and the corresponding ground truth values from MoDOT for two repetitions (Run1 and Run2). The plots depict IRI values for each 0.1-mile section of a 13.7-mile segment on I-70 EB with TWAY ID 19 (Log 133.5 to Log 208.2), with the distance in miles on the horizontal axis and the IRI in inches per mile (in/mi) on the vertical axis. It can be observed that the predictions from our proposed device closely align with MoDOT's ground truth IRI.

To evaluate the performance of our device, we assessed its ability to accurately classify pavement roughness based on the Pavement Ride Quality classification by the U.S. Department of Transportation. This classification scheme was also employed in a similar study conducted by (*4*). The results indicate that our device successfully assigned the appropriate ride quality classes to the pavement segments for both Run1 and Run2. The accuracy of our device in performing this classification task is depicted in the pie charts in **Figure 13** (a) and (b). The accuracy is reported as 97.12% for Run1 and 96.4% for Run2 on I-70 EB. However, it is important to note a marked deviation in the prediction from our device compared to the ground truth at index 32 (Log 136.5 to 136.6). This deviation led to a misclassification of the ride quality and pavement condition in this specific segment. To investigate the cause of this discrepancy, we examined the pavement imagery captured during data collection along I-70 EB. It was observed that this particular road segment (Log 136.5 to 136.6) did not exhibit significant differences compared to the neighboring segments at index 31 (Log 136.4 to 136.5) and index 33 (Log 136.6 to 136.7), which had lower IRI values, as shown in **Figure 14** (a) through (c). The verification process confirmed that the prediction from our device was in good agreement with the observed imagery for this segment.

**Comparing Proposed Device Predictions with Ground Truth IRI for South Providence**
**Figures 12** (c) and (d) present the IRI predictions generated by our device and the actual IRI values provided by MoDOT for the South Providence with TWAY ID 7259 (Log 2.2 to Log 5.5) in two repetitions, namely Run1 and Run2. The results indicate that our proposed device successfully captures some of the patterns in the actual IRI, particularly when the IRI values are below 94 inches/mile. However, our device struggles to accurately predict the IRI when it exceeds 170 inches/mile. This inconsistency can be attributed to the imbalanced distribution of the training data. The training dataset primarily consists of data instances from the I-70EB and I-70WB routes, which constitute approximately 95% of the dataset. These routes exhibit relatively low IRI values (ranging from 20.96 to 80 inches/mile) and high speeds (65 to 70 mph). Consequently, these data instances dominate the overall dataset, as depicted in the histogram plot shown in **Figure 13** (e). Furthermore, the performance of our device was evaluated based on its ability to accurately rate the ride quality of pavement segments. The pie charts in **Figure 13** (c) and (d) demonstrate the device's accuracy in fulfilling this task. For Run 1 and Run 2, the device achieved accuracy rates of 61.29% and 65%, respectively.

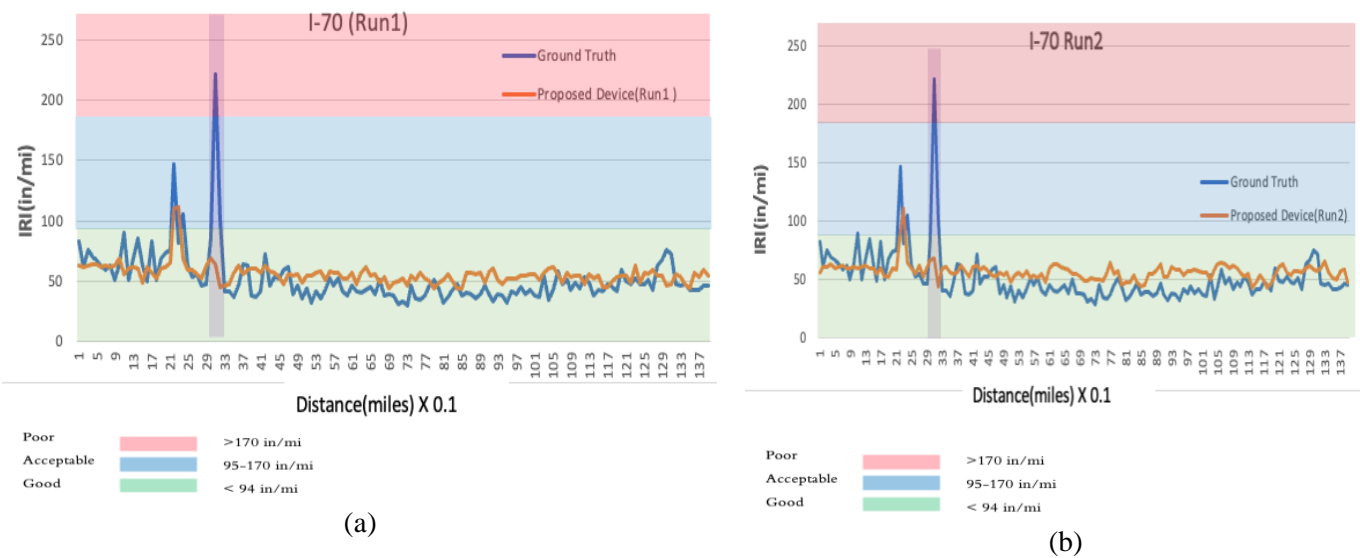

(a)

(b)

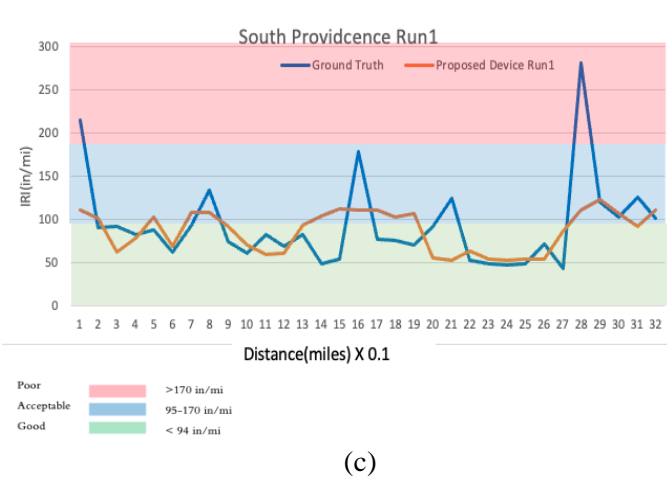

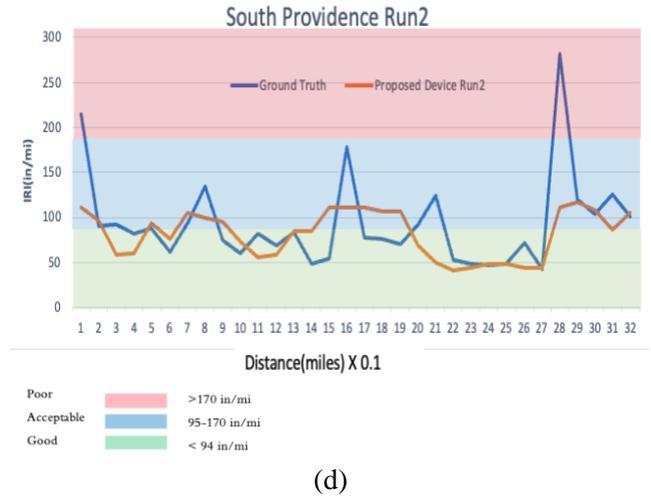

(c)

(d)

**Figure 12 Line Plots of Proposed Device IRI predictions and Ground truth IRI (a) I-70 Run 1 (b) I-70 Run 2 (c) South Providence Run 1 (d) South Providence Run 2**

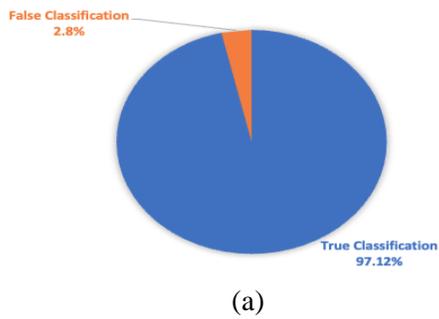

(a)

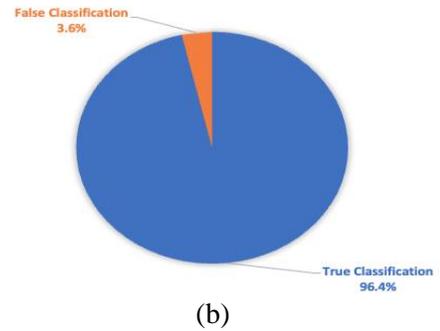

(b)

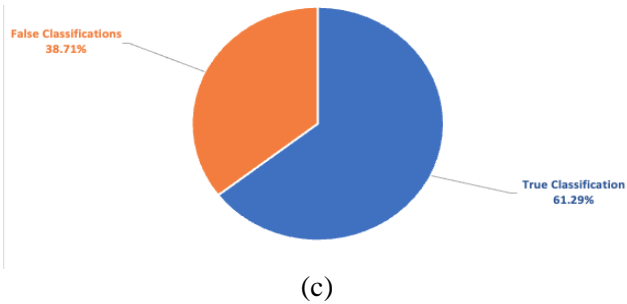

(c)

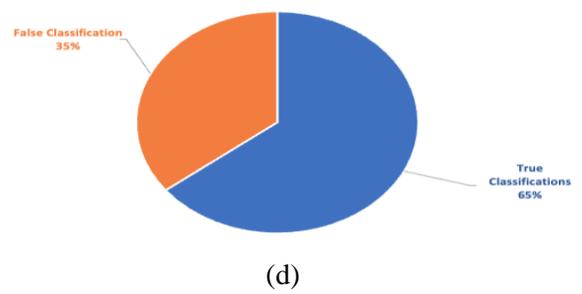

(d)

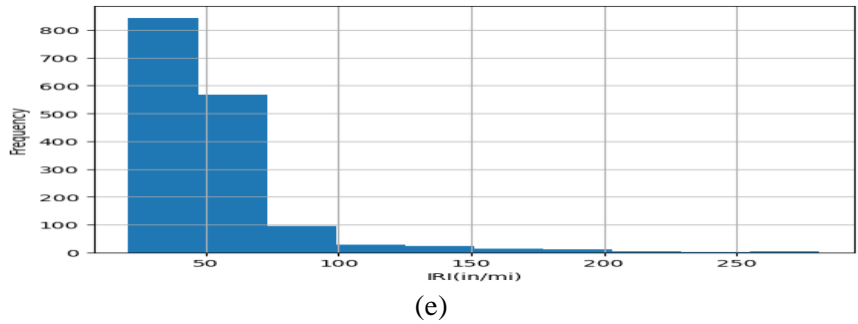

(e)

**Figure 13 Pie Chart of Our Proposed Device's Classification accuracy (a) I-70 Run 1 (b) I-70 Run 2 (c) South Providence Run 1 (d) South Providence Run 2 (e) Histogram of IRI distribution**

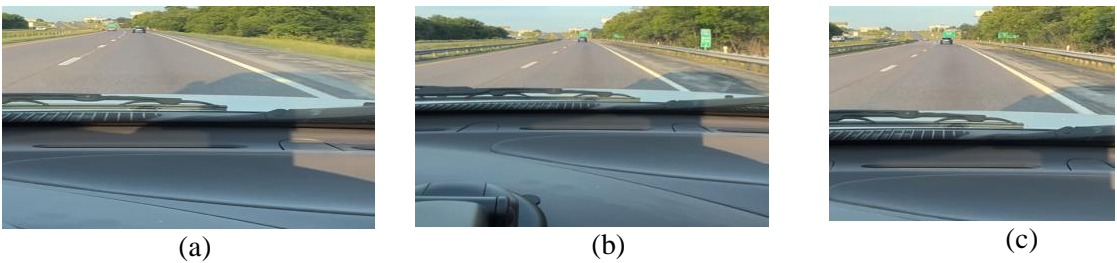

     (a)                     (b)                    (c)

**Figure 14 Images of 1-70EB segment to Verify Discrepancy in IRI Prediction from Device and Ground Truth IRI for Log 136.5 to 136.6: (a) Prior to Segment of interest(b) At the Segment of Interest (c) After the Segment of Interest**

**Repeatability of Proposed Device**

**Figure 15**(a) illustrates the IRI data collected at each 0.1-mile section of the South Providence route, with four repetitions conducted to assess the repeatability of our proposed device. The horizontal axis represents the distance along the driving lane, while the vertical axis represents the IRI values recorded by our device. The chart reveals a high level of consistency among the IRI data from all four repetitions. To quantify these observations, a statistical analysis was conducted on the collected data. In **Figure 15**(b), a plot of the standard deviation (SD) and the Coefficient of Variance (CV) for each 0.1-mile section of the testing route is shown. The average CV for this route is calculated to be 8.89%. According to (*6*), most inertia profiles exhibit a CV of less than 5% for IRI measurements. Considering the significant cost associated with acquiring and maintaining these inertia profilers, an average CV of 8.3% achieved by our portable, low-cost embedded device is considered reasonable. It is important to note that the peak CV for this route reaches approximately 31%. This is expected due to the vehicle wander effect, as discussed by (*6*) making it challenging to drive the survey vehicle along the exact same path for all repetitions.

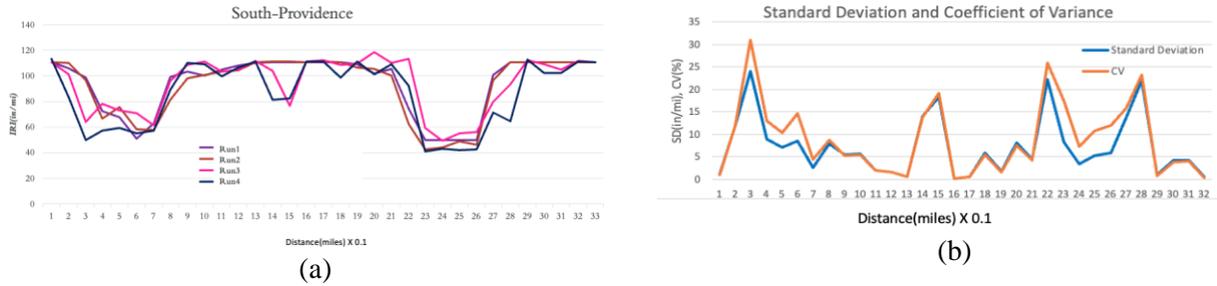

(a)                                                                                                         (b)

**Figure 15 (a) Line Plots of Predicted IRI showing repeatability of Proposed device in 4 Runs (b) Line Plot of SD and Coefficient of Variance for Repeatability Analysis for South Providence Route**

**Comparing Proposed Device with Smart Phone Prediction**

**Figures 16** (a) and (b) shows the line plots of the IRI predictions obtained from our proposed device and the IRI values derived from Smartphone data for the South Providence and Route-K, respectively. A notable similarity can be observed between the predictions generated by our proposed device and those obtained from the Smartphone as they both follow a similar trend. However, it is important to note that our proposed device generally tends to yield higher IRI values compared to the Smartphone. This disparity can be attributed to the positional effect of each data collection device. Our proposed device is mounted on the axles of the data collection vehicle, where the induced vibration from road roughness is most prominent. In contrast, the Smartphone device is positioned in a phone holder attached to the dashboard of the data collection vehicle, as illustrated in **Figure 6** (b). This positioning may dampen the vibration captured by the Smartphone, hence the lower IRI predictions.

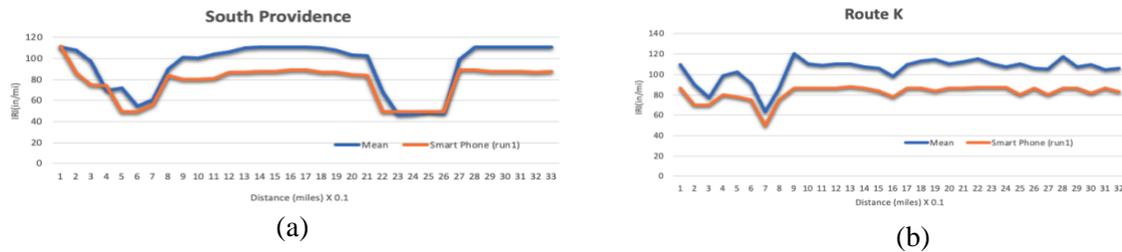

(a)                                                                                                         (b)

**Figure 16 (a) Line Plot of Proposed Device IRI vs Smartphone IRI (South Providence) (b) Line Plot of Proposed Device IRI vs Smartphone IRI (Route-K)**

**CONCLUSIONS**

The strategic timing and prioritization of pavement evaluation and maintenance have assumed unparalleled importance given the challenges of rapidly deteriorating pavement distress, increasing traffic volume and loadings, and constrained transportation funding. The acquisition of Pavement Condition Data, a critical factor in formulating decisions related to pavement maintenance and rehabilitation, has traditionally placed significant financial burdens on state agencies and DOTs. To address this problem, this study introduces a cost-effective system based on IOT and edge computing, designed for collecting pavement roughness data and providing real-time prediction of the IRI. An FFT-based feature extraction algorithm was developed to convert vertical acceleration data from the time domain to frequency domain and extract key spectral features which are used to train three ensemble machine learning models for predicting IRI at every 0.1-mile segment. The feature extraction algorithm and trained machine learning model were deployed on the proposed edge computing device for real time IRI predictions on selected routes (I-70EB, I-70WB, South Providence, Route-K). The results from the device were compared to an industry standard ARAN van from MoDOT and a smart phone-based pavement roughness capture application. The repeatability of this device was also accessed. Overall, it was found that the IRI values measured by our proposed device and that from MoDOTs ARAN van are in good agreement for good-fair conditioned pavement segments.

The conclusion drawn from this study are as follows;

- The comparative analysis of all the three machine learning models shows that XGBoost has the best performance with an RMSE of 16.69in/mi and MAPE of 20.3%. This is followed by the Random Forest Model with an RMSE and MAPE of 16.96in/mi and 22.23%, respectively, on the validation data.
- There is a very good correlation between the IRI predictions from our proposed device and that from the Smart Phone-based data collection, with an average $R^2$ of 0.86. It was also discovered that our proposed device generally gives higher IRI predictions compared to the smartphone, this can be attributed to positional effects as our proposed device is closer to the source of vibration (wheel axle) compared to the smart phone which is mounted on the dashboard.
- The repeatability analysis of our proposed device demonstrates acceptable results when compared to similar studies (*6, 20*). For the South Providence route, the coefficient of variance (COV) ranges from 0.16% to 30.99%, with an average COV of 8.3%. Only three out of the 32 test sections have a COV higher than 20%.

This study has developed a cost effective IoT and edge computing-based device for real time road condition monitoring. This study can be improved in the future by incorporating audiovisual data. This may involve using a more powerful edge computing device such as Nvidia Jetson that can perform real-time audio-visual analytics on such datasets. Also, further studies could collect more data that are more representative of the non-Interstate highways, this may improve the accuracy and the robustness of the device in correctly predicting IRI across multiple routes.

## AUTHOR CONTRIBUTIONS

The authors confirm contribution to the paper as follows: study conception and design: Daud, Adu-Gyamfi, Amo-Boateng; data collection: Daud, Owor; analysis and interpretation of results: Daud, Adu-Gyamfi; draft manuscript preparation: Daud, Aboah, Adu-Gyamfi. All authors reviewed the results and approved the final version of the manuscript.